\documentclass[conference]{IEEEtran}
\usepackage{blindtext, graphicx}
\usepackage[utf8x]{inputenc}
\usepackage{amsmath}
\usepackage{amssymb}

\usepackage{color}

\begin{document}
%
\title{LSTM Networks for Data-Aware Remaining Time Prediction of Business Process Instances}

\author{

\IEEEauthorblockN{Nicolò Navarin, Beatrice Vincenzi, Mirko Polato and Alessandro Sperduti}
\IEEEauthorblockA{Department of Mathematics\\
University of Padova\\
via trieste 63, I-35124, Padova, Italy\\
Email: \{nnavarin, vincenzi, mpolato, sperduti\}@math.unipd.it}
}


\maketitle

\begin{abstract}
Predicting the completion time of business process instances would be a very helpful aid when managing processes under service level agreement constraints. The ability to know in advance the trend of running process instances would allow business managers to react in time, in order to prevent delays or undesirable situations. However, making such accurate forecasts is not easy: many factors may influence the required time to complete a process instance.
In this paper, we propose an approach based on deep \textit{Recurrent Neural Networks} (specifically LSTMs) that is able to exploit arbitrary information associated to single events, in order to produce an as-accurate-as-possible prediction of the completion time of running instances.
Experiments on real-world datasets confirm the quality of our proposal.

\end{abstract}


%
\IEEEpeerreviewmaketitle

\section{Introduction}
{A business process is a collection of activities that takes one or more inputs and creates an output that is of value to the customer~\cite{Burattin2015,Hammer1993}.}
The execution of a process leaves in the \textit{information system} a trace of the performed activities, thus is usually referred as business process instance, or simply \textit{trace}.\\
The prediction of the remaining time of a business process instance is a task that is receiving increasing attention in the last few years.
It has several Business applications, such as in ticketing services, where it is important to give to customers an as accurate as possible prediction of when the problem will be resolved.\\
Several approaches have been proposed to predict the remaining time of activities: \cite{Aalst2011, Tax2017}.
In real world \textit{process-aware information systems} (PAISs), traces and single activities can be associated with side information, such as the person that is performing such activity, the start and end time of such activity, the day of the week and so on.
These additional information can be valuable when predicting the completion time of a trace.
The most basic example is when a process instance, that usually requires one work-day to complete, is started on Friday evening, and the company is closed in the weekend.
In such case, for a human it is immediate to infer that the activity will not be completed until Monday. However, for a predictive algorithm, this information has to be encoded in some way. 
On this simple example, one possibility is to add an if-then-else condition before the prediction, that considers this scenario and, in that case, adds two days (Saturday and Sunday) to the prediction.
However, we are interested in less trivial analyses.
In a general case the needed time to complete a process instance depends on some variables. A a recently approach \cite{Polato2014} proposes to attach to each trace and to each activity, all the available side-information in the form of an attribute-value list (or, equivalently, a vector).

In this paper, we take inspiration from the approaches in \cite{Tax2017} and \cite{Polato2014}.
Namely, we propose to adopt a Long Short-Term Memory (LSTM) network, that is able to handle data attributes associated to the activities.
Differently from \cite{Tax2017}, that adopts multi-task learning, our proposed LSTM is just trained for predicting the remaining time of a trace. Even if authors in \cite{Tax2017} state that predicting together the next activity and its timestamp via a single model is beneficial, in the paper this is shown to hold only when predicting a single activity and its execution time. Indeed, when predicting the whole remaining time of a trace, the approach in \cite{Tax2017} has problems to deal with repeated activities, leading to worse results compared with other (simpler) methods in literature. Moreover, in \cite{Tax2017} it is not clear whether, for the considered problems, multitask learning improves the predictive performances with respect to learning just one task at a time.
%
%
%
Finally, such an approach is not easily extensible to consider other information attached to traces and events.
Indeed, recent PAISs have the capability to store additional information associated to events.
Such information may be very important for many predictive tasks~\cite{Polato2014}.

In this paper, we propose an LSTM architecture for predicting the completion time of a trace, that can consider such additional information.

Our approach improves~\cite{Tax2017} in the following ways:
\begin{itemize}
\item it is able to effectively deal with traces with repeated activities;
\item it can consider data attributes associated to events;
\item it has lower test times;
\end{itemize}
At the same time, our approach improves \cite{Polato2014} in that
it does not require to build a transition system, that can be very time and memory consuming for some real-world datasets.
Moreover, we experimentally show that our proposed approach has higher predictive performances with respect to existing state-of-the-art predictive algorithms, in many real-world datasets.

As a case study, we consider a new dataset generated from an Italian software company. It contains data of the process underlying the helpdesk activity. In particular, it describes the ticketing management process. \\
A new process instance starts when e new ticket is opened. After that, a company resource defines a severity level for the ticket. Then the ticket is assigned to a resource, it is managed and processed. During this phase the ticket could be subject to some anomalies, upgrades or technical operations. Each time the ticket changes step, the old activity stops and a new one is registered, so the current instance continues to follow a path described by the ticketing management process. Finally, when the problem is solved the ticket gets closed and the process instance arrives at the end.\\
The process consists of 10 activities and the event log has been filtered from the original log exported from the company's PAIS. This log has $4,454$ cases with $15,682$ events in total, and the number of attributes associated to events are $8$.


The paper is organized as follows.
Section~\ref{sec:back} introduces some definitions and notation.
Section~\ref{sec:rnn} discusses Recurrent Neural Networks.
In Section~\ref{sec:probstatement} we formally define the problem we face, and in Section~\ref{sec:proposal} we discuss our proposed solution. Section~\ref{sec:related} discusses the related works. In Section~\ref{sec:exps} we experimentally evaluate our proposed method against other state-of-the-art methods in literature. Section~\ref{sec:concl} concludes the paper.

\section{Definitions and notation}
\label{sec:back}

A Business Process (BP) is a collection of related business activities which finds its end in the delivery of a service or product to a customer. 
It is usually represented as a workflow, e.g. with the standard Business Process Model and Notation (BPMN)~\cite{ObjectManagementGroupOMG2011}.
The same BP can generate a different sequence of activities every time it is executed.
A specific execution of an activity is referred to as \textit{event}; a specific execution of the entire process, which corresponds to a path in the workflow model, is referred to as \textit{process instance}.

Formally, an event $e$ is a tuple $e = (a,c,\tau,D)$ where: $a \in \mathcal{A}$ is the name of the process activity associated to the event $e$ (i.e., which tasks was executed); 
$c \in \mathcal{C}$ is the ``case-id'', that is an identifier of the business process instance; 
$\tau \in T$ is a timestamp which indicates the execution time of the specific activity; 
$D \equiv \{(d_1, v_1), \dots, (d_m, v_m)\}$ is a set of attribute-value pairs associated with the execution of the event $e$. 
We assume the presence of a projection operator $\pi$, which allows the extraction of specific attributes out of an event. 
Specifically, given an event $e = (a,c,\tau,D)$, we define $\pi_{\mathcal{A}}(e) = a$, $\pi_{\mathcal{C}}(e) = c$, $\pi_T(e) = \tau$, and $\pi_{d_i}(e) = v_i$.
The event universe $\mathcal{E}$ indicates the set of all possible events.
Moreover, we refer as $\mathcal{E}^*$  the set of all possible sequences over $\mathcal{E}$.
A \textit{trace} $t = \langle e_1,\ldots,e_n \rangle \in \mathcal{E}^*$, of length $|t|=n$, is any of those sequences, where $\forall \: 1 \leq i \leq |t|, \pi_\mathcal{C}(e_i) = c$. A trace is also referred as \textit{business process instance}.

An \textit{event log} over a set of events $\mathcal{E}$ is a set of traces $L \subseteq \mathcal{E}^*$ such that each event appears at most once in the entire log, i.e., for any $t_1$, $t_2 \in L$, $t_1 \neq t_2 $: $\textit{set}(t_1)\cap \textit{set}(t_2) = \emptyset$, where $set(t)$ transforms a sequence $t$ into the set with the same elements.
In the paper we will deal with partial traces. A partial trace can be defined by the \textit{head} operator over sequences $\textit{hd}^k(t)$ as the first $k$ events in the trace, i.e., let $t = \langle e_1, \ldots, e_k,e_{k+1}, \ldots, e_n\rangle$, then $\textit{hd}^k(t) = \langle e_1, \ldots, e_k\rangle$.
The \textit{continuation} of a trace is given by the \textit{tail} operator $\textit{tail}^k(t)$ that returns the last $k$ events in the trace, i.e., $\textit{tail}^k(t)=\langle e_{n-k+1}, e_{n-k+2}, \ldots, e_n\rangle$.

\section{Recurrent networks}
\label{sec:rnn}
When the input depends on time, e.g. when we have to analyze sequences, one of the most known approaches are Recurrent Neural Networks (RNNs).
A recurrent artificial neuron (or unit) is a neuron with a feedback connection (a loop). 
Recurrent neural networks can store representations of past events through those connections.
Indeed, RNNs can be \textit{unrolled}, i.e. they are equivalent to multiple copies of the same network, each passing a message (generated from the past inputs) to a successor (that processes the current input).
More formally, a RNN unit takes in input its previous hidden state and the current input, and outputs a new hidden state.
Let us consider a layer of recurrent units.
In more detail, let $x_t$ be the input of the network at time t, and $h_{t-1}$ be the vector of hidden states for all the recurrent units at time $t-1$. Then we can compute the hidden state at time $t$ as:
\begin{equation}
h_t=\sigma(\mathbf{W}x_t + \mathbf{U} h_{t-1} + b),
\label{eq:rnn}
\end{equation}
where $\sigma$ is the sigmoid function (or other nonlinear functions like hyperbolic tangent or the rectified linear function) applied element-wise, and $\mathbf{W}$, $\mathbf{U}$ are the weight matrices and $b$ the biases (the parameters to learn).
Then the output at time $t$ can be computed as a function of $h_t$.

One can think as this mechanism as a "short-term memory".
It is known that a network composed by several RNN units is Turing complete and can therefore, in principle, implement any algorithm~\cite{Seigelmann1995}.
The main problem of these networks concerns the learning algorithm. The most adopted one is \textit{back-propagation through time}~\cite{Hochreiter1997}, and shows the well-known \textit{vanishing} gradient and \textit{exploding} gradient problems.
These problems make the learning of long-time dependencies very difficult for (vanilla) RNNs~\cite{Bengio1994}. While some values of the parameters able to deal with such sequences exist (i.e. a human could carefully pick the model parameters in order to solve problems with long-time dependencies), in practice learning algorithms seem not to be able to find them.
Long Short-Term Memory~\cite{Hochreiter1997} are special kinds of recurrent neural networks that, coupled with appropriate learning algorithms, can alleviate these problems.

\subsection{LSTM networks}
Long Short-Term Memory networks (LSTM)~\cite{Hochreiter1997,Graves2008} 
are a special kind of RNN, capable of learning long-term dependencies.

\noindent The idea is that, instead of having a recurrent neuron, we have a recurrent \textit{module}.
In particular, this module will have a memory cell that can store information.
An LSTM unit takes in input its old cell state and its old hidden state, and outputs its new cell state and its new hidden state.
More formally, an LSTM unit is composed by four gates, interacting in a very special way.

We can consider all gates of the same kind in a layer of LSTM units together using the following vector notation:
\begin{itemize}
\item forget gate layer: $f_t$;
\item input gate layer: $i_t$;
\item cell state candidate layer: $\tilde{c_t}$;
\item output candidate layer: $o_t$.
\end{itemize}

\begin{figure}
{
    \centering
    \includegraphics[width=\linewidth]{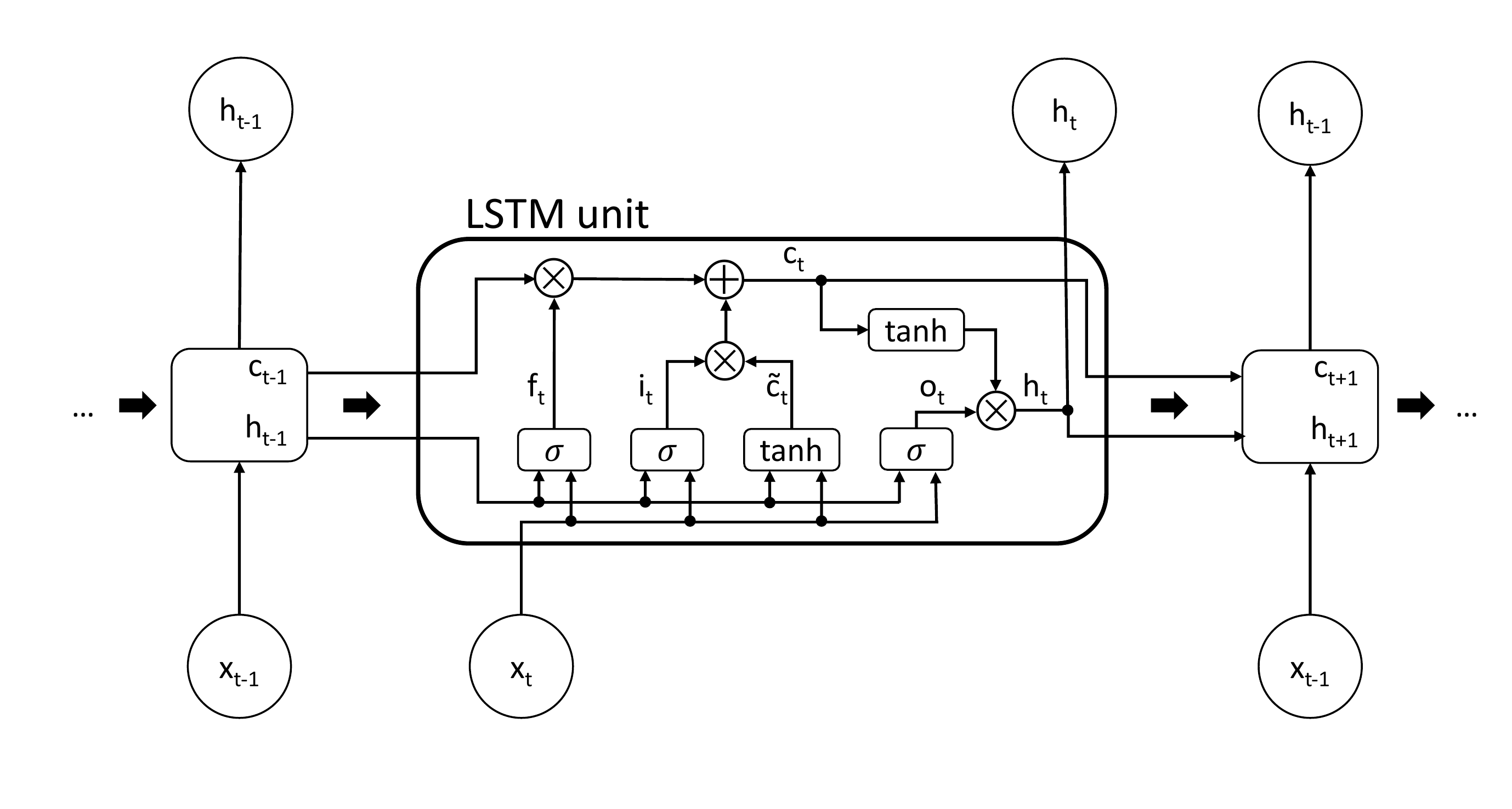}
    \caption{Architecture of a LSTM unit and its evolution through time. The thick arrows indicate the evolution of the unit through time. Black dots indicate connections. }}
    \label{fig:LSTM}
\end{figure}

Let us now detail how these layers interact. Let $x_t$ be the input vector at time $t$, and let $h_{t-1}$ and $c_{t-1}$ be the output and the cell state vectors (respectively) of the LSTM layer at time $t-1$.
Then we can define the output vectors $h_{t}$ and $c_{t}$ of a layer of LSTM cells at time $t$ via the following equations:
\begin{align}
f_t=&\sigma( \mathbf{W}_f x_t + \mathbf{U}_f h_{t-1} + b_f)\\
i_t=&\sigma(\mathbf{W}_i x_t + \mathbf{U}_i h_{t-1} + b_i) \label{eq:i}\\
o_t=&\sigma(\mathbf{W}_o x_t + \mathbf{U}_o h_{t-1} + b_o)\\
\tilde{c_t}=& \textrm{tanh}(\mathbf{W}_{\tilde{c}} x_t + \mathbf{U}_{\tilde{c}} h_{t-1} + b_c)\\
c_t=&f_t \circ c_{t-1} +i_t \circ \tilde{c_t} \label{eq:c} \\
h_t=& o_t \circ \textrm{tanh}(c_t)
\end{align}
where $\circ$ is the Hadamard (element-wise) product, $\sigma$ is the element-wise sigmoid function, $tanh$ the element-wise hyperbolic tangent function, and $\mathbf{W}$, $\mathbf{U}$ and $b$ are the matrices of the parameters and the vector of biases (one corresponding to each gate type).
The outputs of the units are the the new cell state vector $c_t$ (at time $t$), and the hidden state vector $h_t$.
{The LSTM unit architecture and its evolution through time is depicted in Figure~\ref{fig:LSTM}. At time step $t$, the LSTM output $h_t$ depends on the input at that time $x_t$, and on the previous cell states $c_{t-1}$ and $h_{t-1}$.}

A common variation to the original architecture, that allows faster run-times on GPUs, is to couple forget and input gates. In particular, we do not need to compute $i_t$ of eq.~\eqref{eq:i} anymore, and eq.~\eqref{eq:c} becomes:
\begin{equation}
c_t=f_t \circ c_{t-1} + (1- f_t) \circ \tilde{c_t}.
\end{equation}
In this paper, we adopt this variation.

Let us draw some considerations on how the LSTM can alleviate the vanishing/exploding gradient problems when trained with back-propagation through time.
When we compute the gradients of the error function during backpropagation, their magnitude is influenced by two factors: the weights and the derivatives of the activation functions.
If either one of these factors is smaller than 1, then the gradients may vanish in time; if larger than 1, then they may explode.

Let us consider the cell state $c_t$ of eq.~\eqref{eq:c}. The second part of the right-hand side of the recurrency (after the $+$ sign) is some function of the inputs. When we backpropagate and take the derivative of $c_t$ with respect to $c_{t-1}$, this added term disappears. In the first part of the recurrency, the cell state is multiplied  by the output of the forget gate, so we can see $f_t$ as the weights for the cell state. In this case, there is no activation function (besides the identity, which derivative is always one). 
So, if the forget gate is on (activation close to 1.0), then the gradient does not vanish. Since the forget gate activation is never greater than 1.0, the gradient can't explode either.
This does not solve the vanishing gradient problem in general, but improves the situation a lot compared to vanilla RNNs of eq.~\eqref{eq:rnn}.

\section{Problem statement}
\label{sec:probstatement}
The problem we face in this paper is the remaining time prediction of business process instances.
Let us assume there is a Business Process (or a set of them) that generates some business process instances, i.e. traces.
Given a trace $t=\langle e_1,\ldots, e_n \rangle$, it is straightforward to compute the amount of time that has been required for its execution: we just have to subtract the timestamp of the first event in the trace from the one of the last event. We refer to this quantity as the execution time of a trace $et(t)=\pi_T(e_n)-\pi_T(e_1)$.

When a process is still running (i.e. it has been not completed yet), we may be interested in estimating how much time is left for it to be completed. In this scenario, a (possibly small) part of its events have been performed. Our problem is then to estimate the difference between the (unknown) process completion time, and the time that passed since its first event has started. More formally, given a trace $t=\langle e_1,\ldots, e_n\rangle$ of which only the first $k$ events are known (i.e. we are given $hd^k(t)$), we want to learn a function $f(\textit{hd}^k(t))$ that predicts $et(t)-et(\textit{hd}^k(t))=et(\textit{tail}^{n-k}(t))$.

Note that in this paper we are interested in predicting just the remaining time for a trace, and not the exact activities that will happen. Indeed, in many real world applications, we are just interested in the completion time of a trace.
As a motivation case study, we consider the helpdesk sector of an Italian software company. The remaining time information can be communicated by a helpdesk resource to the customer. In this case, the customer is not interested in the exact steps that will happen, she just wants to know how long she has to wait for her ticket to be solved, as accurately as possible. 
On the other hand,  the  ability  to  know in  advance  the  trend  of  running  process  instances  would  allow business managers to react in time, in order to prevent delays or undesirable situations.

\section{Related works}
\label{sec:related}
The literature proposes plenty of works aiming at improving business processes and provide support for their execution. Historically, the first approaches were based on measuring and monitoring activities and belong to the research area called ``Business Activity Monitoring'' \cite{Golfarelli2004}.
The introduction of process mining techniques \cite{VanderAalst2011a} allowed the automatic analysis of running process instances and the prediction of different aspects for them.

One of the first works that analyzes the execution duration problem is~\cite{Reijers2006}. This particular work concentrates on cross-trained resources, but no detailed prediction algorithm is reported.
In \cite{VanDongen2008,Crooy2008}, van Dongen et al. describe a prediction model which uses all the data recorded in an event log. This approach uses non-parametric regression in order to predict the ``cycle time'' (i.e., the remaining time) of running process instances.
The recommendation system, described by van der Aalst et al. in \cite{Schonenberg2008}, is built using historical information, and is able to predict the most likely activity that a running case is going to perform.
The TIBCO Staffware iProcess Suite \cite{Schellekens2009} is one of the first commercial tools that predicts the remaining time of running process instances. This tool simulates the complete process instance without analyzing historical data. The main building blocks of the prediction are parameters, provided by the users at ``build time'', such as the process routing or the expected duration of activities.
The first successful framework focused on the time perspective is proposed in \cite{VanderAalst2009}. Song et al. describe an approach which builds a finite state machine, called transition system, with respect to a given abstraction of the events inside the events log. An abstraction is a function which maps a trace to a state and it can be very influential to the success of the method. These functions generally create a mapping from a sequence of events (i.e., the trace) to an easier entity, e.g., the set of its events' activity names. These mappings depend on an hyperparameter (the horizon) that is proportional to the dimension of the state space.
The most popular abstractions are: the set, the bag (also called multiset) and the sequence. 
After the construction of the transition system is then augmented with time information about the historical process instances. The time forecast is performed using statistics over the time information (e.g., mean duration of specific class of cases) collected inside the state of the transition system corresponding to the current running instance.  We refer to this method as \textit{VDA} in Section\ref{sec:exps}. The suffixes \textit{-SET}, \textit{-BAG} and \textit{-SEQ} refer to the used abstraction for constructing the transition system.
A similar approach is presented in \cite{CeciLFCM14}. In this work the authors decorate the transition system with decision trees that are trained, over the collected data, and then used to predict the completion time. Moreover, the proposed method is also able to predict the next activity that a process instance is going to take.
One of the first methods which has taken advantage of additional data (i.e., resources) is presented in \cite{Leitner2010}. The approach considers the data perspective in order to identify SLAs (Service Level Agreement) violations. The actual forecast is built using a multilayer perceptron, trained with the Backpropagation algorithm.
An extended version of the technique described in \cite{VanderAalst2009} has been proposed by Folino et al. \cite{Folino2012,Folino2013}. In particular, the method clusters the log traces according to the corresponding "context features" (i.e., the additional data of a trace) and then, for each cluster, a predictive model is created by using the method described in \cite{VanderAalst2009}. Clustering trees are used to form the clusters. At prediction time, the approach associates the new running instance to one of the clusters and then uses the model belonging to that specific cluster to make the prediction. 

A probabilistic approach able to predict the likelihood of future activities, called \emph{Instance-specific Probabilistic Process Models} (PPM), is reported in \cite{Lakshmanan2013}. Even though it does not provide a remaining time prediction, it provides to business managers useful insights regarding the progress of the process.
Another probabilistic method, based on Hidden Markov Models (HMM), is proposed in \cite{Pandey11}. Experimental results seem to show how that the method outperforms the state-of-the-art. However, the achieved results are not much different from the ones achieved by the transition system based method presented in \cite{VanderAalst2009}.
In a recent work \cite{Ghattas2014}, Ghattas et al. exploit, what they call, \emph{Generic Process Models} and decision trees, in order to provide decision criteria defined according to the actual process goals.
%
%
In \cite{Polato2014}, Polato et al. proposed another approach based on \cite{VanderAalst2009} in which the additional attributes of the events are used in order to improve the remaining time prediction quality. This method exploits the concept of annotated transition system (presented in \cite{VanderAalst2009}) by adding machine learning models, such as Na\"ive Bayes and Support Vector Regressor. Experimental results show how the additional attributes can positively influence the prediction quality. We refer to this method as \textit{DATS} in Section \ref{sec:exps}.

There are also approaches coming from different areas. For example, queue theory~\cite{hall1990queueing,bolch2006queueing} and queue mining can be seen as an instance of process mining, and recent works are starting to aim for prediction purposes~\cite{Senderovich2015278}. Here authors focus on the delay prediction problem, i.e., providing information to user waiting in lines, in order to improve customer satisfaction. The method leverage on an annotated transition system. That model is then used to make delay predictions using simple averages or non-linear regression. 

More recently, deep neural network based approaches have been proposed. In \cite{Evermann2017}, a recurrent neural network (RNN) for the prediction of the next process event. The network is composed by two hidden RNN layers using basic LSTM cells (Long-short term memory). Experimental results have shown good precision on the BPI challenge 2012 dataset.
A closely related task to the remaining time prediction is the predictive monitoring, in which the goal is to provide early advice so that users can steer ongoing process executions towards the achievement of business constraints. In \cite{Maggi2014}, Maggi et al. present a decision tree based predictor model. This framework continuously estimate how likely is that a user defined constraint will be fulfilled by the current process instances. The same task is faced in \cite{Leontjeva2015}, where authors present an approach in which traces are treated as complex symbolic sequences. They propose different possible encodings: \textit{(i)} index-based and \textit{(ii)} a combination of the first one and an HMM based one. From an AUC (Area Under the Roc Curve) perspective the proposed encoding outperforms the baselines.
Using the same encoding, Verenich et al. \cite{Verenich16} proposed a two phases approach: in the first phase the dataset is split in groups by means of unsupervised clustering; in the second phase a model is trained for each cluster. They used random forests as the predictor algorithm. A very similar approach is presented in \cite{dfm2017}.


The closest related work to the one presented in this paper is~\cite{Tax2017},
where the authors propose a multitask learning modeling based on LSTM.
The model predicts the next activity and its duration.
Then, they are able to predict the continuation of a trace by iteratively predicting the next activity, until the end of case is predicted.
This approach shown bad performance when the log has many repeated events.
%
%
In~\cite{Tax2017}, the comparison with simpler LSTMs trained for the single tasks is presented just in the setting of the prediction of the single next activity and its completion time. Such comparison in the task of predicting the completion time of a trace is missing.

\section{Our Proposal}
\label{sec:proposal} 
We borrow from~\cite{Tax2017} the idea of applying LSTM on the trace remaining time prediction problem.
Indeed, as detailed in Section~\ref{sec:back}, a trace is modeled as a sequence of events, which is exactly the type of input a LSTM expects.
Moreover, in recent \textit{Process Aware Information Systems}, events can have information associated to them, that can be precious in predictive tasks such the one we are considering~\cite{Polato2014}.
In our modeling (see Section~\ref{sec:back}), we assume such additional information to be encoded in a fixed-size vector.
Such vector can have both categorical or real-valued feature.
Given an event log $L$,
we have to define a representation that is suitable for an LSTM network.

LSTM is defined over sequences of vectors as inputs, and hence we need to encode the event and its attributes in a fixed-size vector.
To do that, we encode the activity of the event ($\pi_\mathcal{A}(e)$) using a \textit{one-hot} encoding. 
Moreover, we compute other features for each event, such as the time from trace start, time from last event, the time and weekday in which the event started.
Finally, we append to these features the encoded event attributes, in which categorical attributes are again one-hot encoded, generated according to the following procedure.\\
\noindent Let us  assume we know the number $m$ of attributes.
Let us also assume to have a function $\textit{count}_{|d_i|}(L)$ that returns the number of distinct values that $d_i$ can take in the log $L$ if $d_i$ is a categorical variable, or $1$ if $d_i$ is a real-valued attribute.\\
Then we can define the vector $a$ of length $\sum_{i=0}^{m} {count}_{|d_i|}(L)$ that will be our vectorial encoding for the attributes associated to an event $e$. For each attribute $d_i$, we will set $a[\sum_{j=0}^{i-1}{count}_{|d_j|}(L)]=v_i$ if it is a real-valued attribute, or $a[\sum_{j=0}^{i-1}{count}_{|d_j|}(L)+h^i(v_i)]=1$ if it is categorical. Note that we use the \textit{hashing trick}, with he function $h^i(v_i)$ that maps from the space of possible distinct values of $d_i$ to the set of integers $\{0,\ldots,count_{|d_i|}(L)\}$.\\
Eventually, we end up with a vector representation of an event, which size depends on the dataset, but that can be determined a-priori knowing which data is available.

At this point, a trace is a sequence of events in their vectorial form. We can fix the maximum length of a trace (e.g., the maximum length of all the traces in the training data) obtaining a matrix representation for each (possibly partial) trace, where the matrix is filled with padding rows if the length of the trace is lower than the number of rows in the matrix.
This matrix is a training example for our LSTM.

Note that, for training, we excluded completed traces that would have target to zero since it is out of the scope of our considered problem to predict if a trace is completed or not.

\section{Experimental Results}
\label{sec:exps}
In this section, we compare our proposed DA-LSTM method against LSTM~\cite{Tax2017}, three variants of DATS~\cite{Polato2014} and three variants of VDA~\cite{Aalst2011} on three real-world datasets.
\subsection{Dataset description}
For our experimental we adopted two real world datasets, one of which have been filtered to simplify the task, keeping only the traces with a certain endpoint activity.
The result of this processing is a third, simpler dataset, that however reflects real world data.
In the following, we describe in detail each dataset.
\begin{itemize}
\item \textbf{Helpdesk2017}: this log is a real-life log of SIAV s.p.a. company in Italy. The event log represents instances of a ticketing process in the company helpdesk area. The cases range from July 2016 to February 2017. The log has 15,682 events, 4,454 cases and 10 activities in total. Each event has 7 attributes: \textit{resource}, \textit{customer}, \textit{product} are literal type instead \textit{severity}, \textit{serviceLevel}, \textit{serviceType}, \textit{ticketDifficulty} are numeric type.
\item \textbf{BPI12}: this is a filtered version of the dataset from BPI 2012 challenge. The BPI 2012 challenge dataset represents a real-life log of Dutch Financial Institute. The event log describes an application process for a personal loan or overdraft within a global financing organization. The data has been filtered, we have taken events that are performed manually. Further, we have considered only events of type \textit{complete}. 
This is the same dataset as in~\cite{Tax2017}: it contains 9,658 cases, 72,413 events and 6 different activities. There are two different event's attributes the \textit{amount\_req}, that represents the amount requested by the customer. It's a global attribute, i.e. every case contains \textit{amount\_req} attribute and it never changes in the same trace. The other one is the \textit{resource} of the activity and the set has 60 values.
\item \textbf{BPI12\_oneEndAct}: this is a subset of BPI12 dataset. In the filtering we have considered only traces with the same endpoint activity. The selected event is \textit{W\_Valideren aanvraag-COMPLETE}, because many traces finish with this activity. The final log has 31,829 events, 2,751 cases and 6 activities.
\end{itemize}

\subsection{Implementation details}
We implemented our proposed DA-LSTM network using the \textit{Keras} framework~\cite{chollet2015keras}.
For the LSTM~\cite{Tax2017} baseline, we fixed the number of shared layers to $1$ as suggested in the original paper.
For both our approach and the LSTM~\cite{Tax2017} baseline,
we tested different network configurations, validating the number of LSTM neurons $n$ per layer in $\{100,150,200,250\}$ and the number of layers $l$ in $\{1,\ldots,6\}$.
Note that, for the proposed DA-LSTM, the results are relatively stable varying the architecture (at most approximately 0.025\% difference in MAE between the worst and the best architecture).
As learning algorithm, we adopted Nadam~\cite{Dozat2016}.
The VAD and DATS baseline experiments have been performed in \textit{ProM 6.5.1}, that is an open-source framework of process mining~\cite{VanDongen2005}. It contains many plugins and it is written in Java. In particular we used the plugin implemented in \cite{Polato2014}.
For the VDA baseline, we tested all possible abstractions for transition system construction (\textit{sequence}, \textit{set} and \textit{bag}).
In the same way, for the DATS baseline, we have done a comparison between all abstractions. We used a $6 \times 6$ grid to search the best hyperparameters, where the minimum value is $10^{-3}$ and the maximum value is $10^2$. The considered hyperparameters are the $\gamma $ parameter for RBF kernel and the $C$ cost for SVR. We tested all possible pairs and chose the one with the minimal validation error.

BPI12 dataset is a real-life event log of a financial institute, which has a very complex workflow.
For this reason the transition system construction for \textit{sequence} and \textit{bag} abstractions with the horizon set to infinity is hard. The plugin creates more then $5,000$ states and it is heavy to store and access in memory. So, for \textit{BPI12} and \textit{BPI12\_oneEndAct} datasets, we have decided to reduce the horizon to $5$ for \textit{sequence} and \textit{bag} abstractions.

\subsection{Experimental setup}
As stated in Section~\ref{sec:probstatement}, we face the problem of predicting the remaining time required from a partial trace to complete.
In order to assess the performances of our proposed method, we adopted the datasets presented in the previous section.
The considered datasets comprehends only complete traces.
In order to derive partial traces from a trace $t=\langle e_1, \ldots, e_n \rangle$ of length $n$, we generate one partial trace for each prefix length $1\leq k \leq n-1$, i.e. the set of partial traces corresponding to a trace is $\{hd^k(t) | 1\leq k \leq |t|-1\}$.
Differently from~\cite{Tax2017}, we consider all prefix lengths, including partial traces with just one event. This choice is motivated by the fact that the proposed method, if implemented in a real business environment like the helpdesk of our case-study, has to be able to predict a completion time right after the ticket has been opened. In this way the remaining time predicted can help company resources to inform the final client how long will the ticket takes to be resolved.
We use the first $2/3$ of the traces as training and validation data (we randomly sample the 20\% of training data as validation set).
We use the valdation set for selecting the best hyper-parameters and network architecture.
Finally, we evaluate the time predictions on the remaining $1/3$ of the traces (test set). \\
As performance measure, we consider the \textit{Mean Absolute Error} (MAE), that is an error measure among continuous variables. 
Let $y_i$ the desired output, and $\hat{y_i}$ be the output predicted from a model on $x_i$, for some data samples $\{x_i | 0\leq i\leq n\}$. Than we can define the MAE as:
\begin{equation*}
MAE=\frac{\sum_{i=0}^n |y_i - \hat{y_i}|}{n}.
\end{equation*}
The MAE has an intuitive interpretation as the  average absolute difference between the prediction and the expected output.
Being an error measure, the lower the better.
We computed the \textit{root mean squared error} as well, and the results were concordant so we decided not to report them.

\begin{table}[t]
    \centering
    \begin{tabular}{|l|c|c|c|}
    \hline
   Method/Dataset & Helpdesk2017 & BPI12 & BPI12\_oneEndAct \\
    \hline
    VDA-SEQ~\cite{Aalst2011}  & 5.95 & 8.74~  & 6.60~ \\
    VDA-SET~\cite{Aalst2011}  & 6.03  & 8.45~  & 5.80~  \\
    VDA-BAG~\cite{Aalst2011}  & 6.06  & 8.69~  & 6.58~ \\
    DATS-SEQ~\cite{Polato2014}  & 5.27  & 8.13* &5.44*\\
    DATS-SET~\cite{Polato2014}  & 5.53  & 8.15~ &5.43~ \\
    DATS-BAG~\cite{Polato2014}  & 5.37  & 8.12*  &5.50*\\
    LSTM~\cite{Tax2017} & 4.03  &9.74~ &5.95~ \\
      & (n=150, l=5)  &(n=150, l=5) &(n=100, l=3) \\

\hline
   DA-LSTM  &\textbf{4.01} & \textbf{7.04}& \textbf{4.65} \\
    & (n=250, l=5) &(n=100, l=4) &(n=250, l=1) \\
   \hline
    \end{tabular}
    \caption{Comparison of our proposed method (DA-LSTM) and the considered baselines.
    The results are reported in  MAE (days).  (*): The abstraction horizon is not infinity, but it is fixed to 5.}
    \label{tab:results}
\end{table}

Table~\ref{tab:results} reports the results of our experiments. For each dataset and method, we report the MAE in days.
For the two methods based on LSTM, we furthermore report the number of neurons $n$ and number of layers $l$ of the selected network architecture.
Note that, on this scale, it makes little sense to compare decimal points, i.e. for the results discussion, we will round the results considering whole days.
In the \textit{Helpdesk2017} dataset, LSTM~\cite{Tax2017} is the better performing method among the competitors. Our proposed DA-LSTM shows basically the same performance.

Let us now consider the \textit{BPI12} dataset.
One characteristic of this dataset is that the same trace may have multiple events corresponding to the same activity. This is an issue for LSTM~\cite{Tax2017} (see Section~\ref{sec:related}), that is the worst performing method on this dataset.
The four variations of VDA~\cite{Aalst2011} generally perform slightly better, with the \textit{SET} abstraction being the better performing one.
The DATS~\cite{Polato2014} method shows improved performance compared to VDA. In this case, the BAG abstraction is the best among the three variants.
Our proposed method is the better performing one, with a MAE difference with respect to the best competitor of more than $1$ day.

As for the \textit{BPI12\_OneEndAct} dataset, VDA-SET~\cite{Aalst2011} and LSTM~\cite{Tax2017} show comparable performance, with the former showing a slightly lower MAE.
DATS~\cite{Polato2014} is again the better performing method among the competitors, with all its variants performing better than VDA and LSTM.
In this dataset, our proposed DA-LSTM is able to improve the performance of DATS, showing again the best predictive performances among the considered methods, with a MAE gap of almost $0.8$ days on the second best method.

From the table we can see that the proposed DA-LSTM method has always higher predictive performance with respect to the best baseline approach.

From these results it emerges that for the \textit{Helpdesk2017} dataset the most important information lies in the workflow, and not in the additional events' attributes.
On the contrary, for the \textit{BPI12} dataset, the techniques that consider additional attributes (DATS and the proposed DA-LSTM) are the best performing ones, showing the importance of considering such attributes in the learning procedure.

A note on computational times.
The computational time required for training our proposed model is in general lower compared to \cite{Tax2017}, because we have a single output layer.
The main advantage is in the prediction phase, where our approach have to perform just one prediction per trace while, on the other hand, \cite{Tax2017} have to perform one prediction for every (predicted) activity in the continuation of the trace.
From a real-world point of view, this difference may be important since, for instance in our helpdesk case study, having a fast prediction means that it is possible to give an user the expected completion time right after the insertion of the ticket in the system (e.g. while he is still on the phone), even under heavy load. In this way the customer can know in a more precise way the expected waiting time, and the helpdesk service can monitor the behavior of such predictions in order to maximize its efficiency.

\section{Conclusion}
\label{sec:concl}
In this paper, we proposed a new method for the problem of remaining time prediction of Business Process Instances. Our approach is based on LSTM, a specific type of Recurrent Neural Network. It allows to associate additional information to each event, that may be useful for the task. We demonstrated the quality of our proposal with experiments on several real-world datasets.
As a future work, we plan to explore techniques for alleviating the encoding size resulting from having many categorical features among the attributes. Indeed, when the universe of values a discrete value can take is high, the dimensionality of the vectorial encoding of each activity grows as well. 
Hashing techniques and random projections can be applied in order to make the approach more space-efficient.

Moreover, we plan to apply other machine learning techniques to the problem faced in this paper, e.g. kernels for sequences coupled with SVM.

\bibliographystyle{abbrv}
\bibliography{Mendeley,library-mirko}

\end{document}